\documentclass[journal]{IEEEtran}
\ifCLASSINFOpdf
\else
\fi
\usepackage{microtype}
\usepackage{graphicx}
\usepackage{subfigure}
\usepackage{booktabs} 
\usepackage{multirow,multicol}

\usepackage{hyperref}

\begin{document}
%
\title{Arbitrary-sized Image Training and Residual Kernel Learning: Towards Image Fraud Identification }
%
%
%

\author{Hongyu~Li,~\IEEEmembership{Member,~IEEE,}
        Xiaogang~Huang,~
        Zhihui~Fu,~        
        and~Xiaolin~Li
\thanks{The authors are with AI Institute, Tongdun Technology. (e-mail: hongyu.li@tongdun.net)}}

\maketitle

\begin{abstract}
	Preserving original noise residuals in images are critical to image fraud identification. Since the resizing operation during deep learning will damage the microstructures of image noise residuals, we propose a framework for directly training images of original input scales without resizing. Our arbitrary-sized image training method mainly depends on the pseudo-batch gradient descent (PBGD), which bridges the gap between the input batch and the update batch to assure that model updates can normally run for arbitrary-sized images. 
	In addition, a 3-phase alternate training strategy is designed to learn optimal residual kernels for image fraud identification. With the learnt residual kernels and PBGD, the proposed framework achieved the state-of-the-art results in image fraud identification, especially for images with small tampered regions or unseen images with different tampering distributions. 
\end{abstract}


%
\IEEEpeerreviewmaketitle

 \section{Introduction}
 \label{sec:introduction}
 \begin{figure*}[th]
 	\vskip 0.2in
 	\begin{center}
 		\centerline{\includegraphics[width=0.8\textwidth]{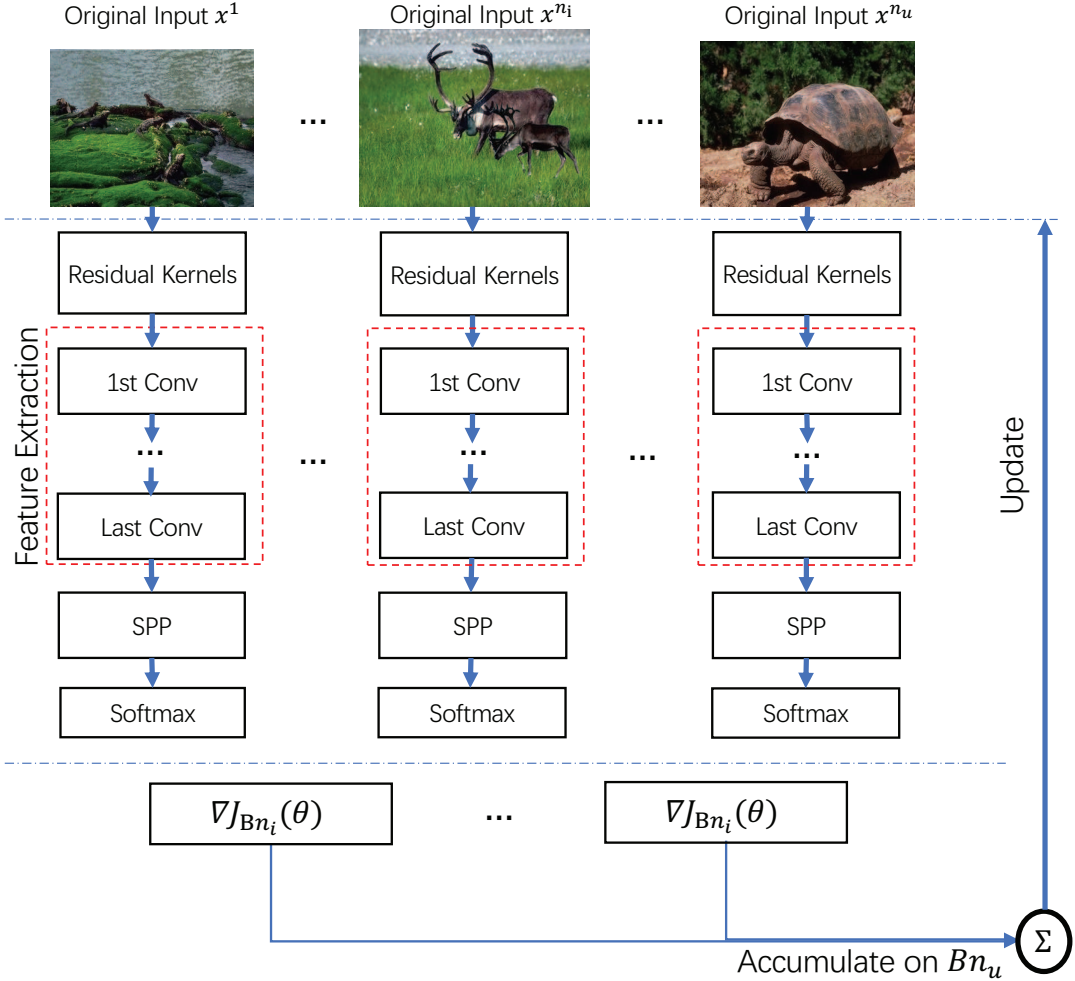}}
 		\caption{The proposed framewrok for image fraud identification. The residual kernel layer is essentially a convolution layer and can be ignored for general-purposed learning tasks. The noise residuals are nonlinearly transformed by the feature extraction layers followed by the SPP layer. With PBGD and SPP, arbitrary-sized images can be fed into the network to accumulate gradients for optimization.}
 		\label{fig:framework}
 	\end{center}
 	\vskip -0.2in
 \end{figure*}
 
 Image fraud is becoming simpler and more accessible with the rapid advancement of the image-editing software. As a result, image fraud identification has attracted much attention of Internet finance companies and the vision community in recent years. Two types of image forgery, copy-move and splicing, are frequent and thus more concerned. For copy-move, pixels are copied and pasted within an image, while splicing generally joins image patches from different images by sticking them together.
 
On copy-move and splicing detection, most methods depend on such intrinsic information as image semantics or image noise residuals. In order to make full use of such information, the methods usually deal with images in two different ways, blockwise or global. The blockwise approaches \cite{li2007sorted,lin2011integrated} extract compact representations of latent features on local image patches. Obviously, they tend to suffer from high false positive rates on fraudulent images and the redundant calculations on the overlapping patches will lead to low training and predicting efficiency. As for the global approaches \cite{rao2016deep,zhou2018learning}, calculations are directly performed on the full image with convolutional neural networks (CNNs). However, the input images are generally required to be enlarged or shrunk to a fixed scale for CNNs. Enlarging will lead to the introduction of new resampling traces in images, and shrinking will discard the details of intrinsic features especially for small tampered areas. Both enlarging and shrinking may deteriorate image fraud identification.

In this paper, we continue moving forward along the research line of global methods and propose a framework for directly training images of original input scales without resizing.
Although spatial pyramid pooling (SPP) \cite{he2014spatial} is the first work proposed to solve the problem of arbitrary-sized image training, only few resolutions are predefined in SPP-net and the images still need to be rescaled to the predefined resolutions. It is also possible to completely bypass the resizing operation if setting the mini-batch size to 1, where the model optimization actually adopts stochastic gradient descent, i.e. SGD. In this case, both batch sizes used to estimate gradients (hereafter referred to as \textit{the input batch}) and to update model parameters (hereafter referred to as \textit{the update batch}) are 1, so the model will converge in a slow and oscillating way.
In the proposed framework, our arbitrary-sized image training method mainly depends on the pseudo-batch gradient descent (PBGD), which bridges the gap between the input batch and the update batch. PBGD accumulates the gradients sequentially and makes updates adaptively. That is to say, the input batch size is a certain number in PBGD, while the update batch size can be another number no less than the input batch size, as shown in Figure~\ref{fig:framework}.

%
%
%
%
%

To adapt our framework for image fraud identification, we also propose a residual kernel learning method to extract
image noise residuals that have specific patterns for both copy-move and splicing forgeries.
Recent research shows that 3 noise filters \cite{zhou2018learning} are sufficient to generate noise residual features. Nevertheless, these filters (i.e. residual kernels) are fixed as preprocessing during learning, which is hard to simulate the complicated and ever-changing noise residual distributions on images. In our work, the residual kernel weights are trained together with the other layers in the network in a 3-phase alternate training fashion. First, the residual kernel layers are frozen and the remaining are trained. Then, freeze the latter while training the former, and alternate between training and freezing. Finally, relax and train both of them simultaneously. 
Although the residual kernel layers usually have weak gradients during back propagation due to their being at the bottom of CNNs, this alternate training strategy can assure that better residual kernels are learnt while preventing vanishing gradients.
The proposed framework is evaluated on several standard datasets for image fraud identification, achieving the state-of-the-art results as well as demonstrating superiority on small tampered regions and strong generalization capability to new dataset.
 
In summary, the contributions of this paper are two folds: 1) the introduction of PBGD in the proposed framework makes it possible to train arbitrarily-sized images, and achieves better performance in image classification through avoiding the semantic category change caused by cropping, scaling or warping. 2) the optimal residual kernels are learnt with a 3-phase alternate training strategy and can significantly improve the accuracy for image fraud identification.
 
 \section{Related Works}
 
This work is mainly related to three research fields, gradient descent, image feature representation, and image forgery detection. This section will briefly introduce related works from the above three aspects.
 
\subsection{Gradient Descent} \label{sec:gd}
 
Gradient descent is widely used to optimize neural networks \cite{sgd2003}, which minimizes an objective function by updating the parameters in the opposite direction of the gradient. 
Gradient descent can vary in terms of the number of training examples used to calculate error and update the model.
The three main variants of gradient descent are batch, stochastic, and mini-batch.
Batch gradient descent (BGD) calculates the error and updates the model on all training examples, but stochastic gradient descent (SGD) does for each example in the training dataset. As a trade-off between SGD and BGD, mini-batch gradient descent (MBGD) splits the training dataset into small batches that are used to calculate error and update model coefficients, which is opposed to the SGD batch size of 1 example, and the BGD size of all training examples.
MBGD seeks to find a balance between the robustness of SGD and the efficiency of BGD \cite{Bottou:2007}. It is the most common implementation of gradient descent used in the field of deep learning.

The structure of tensors in mainstream machine learning platforms, such as TensorFlow, makes the implementation of MBGD more efficient, which requires that images in each batch must be of same size. 
Our work aims to relax this strict requirement to handle arbitrarily varying input scales in the practical implementations. 

 
 \subsection{Image Feature Representation }
Image feature representation with fixed length has been widely studied in computer vision communities. Bag-of-Visual-Words is the basis of
many modern features \cite{BoVW2010} for image recognition and retrieval. It treats image descriptors as visual words, whose counts of occurrence are computed to form the final representation. Varied descriptors are normalized to fixed-length vectors as image feature representation. Spatial Pyramid Matching (SPM) is an extension of the orderless BoVW representation \cite{SPM-2006} by computing histograms inside each subregion. Via introducing the geometric relationship, the performance is significantly improved beyond simple BoVW. SPP based networks \cite{he2014spatial,yue2016deep,qu2017vehicle} are inspired from SPM and aim to convert varied-sized feature maps in CNNs into fixed-length feature vectors while reserving the geometric information. The main disadvantage of these networks is that they can be trained only with few specified resolutions, still causing loss of information. To solve this problem, \cite{li2018spatial} adopted SGD instead of mini-batch SGD during training and conducted studies on a small network. Since large-scale networks such as VGG \cite{Simonyan2014VeryDC} are hard to converge with SGD, we still need to reconstruct the implementation for fixed-length image feature representation.

 \subsection{Image Forgery Detection} 
It has been widely recognized that the key to detect image forgery lies in the low-level image artifacts, including wavelet subbands, low-level image descriptors, color filter array (CFA) and image noise residuals. 

For copy-move forgeries, \cite{li2007sorted} extracts similar singular vectors of low-frequency blocks in wavelet subbands, and \cite{lin2011integrated} instead utilizes SURF \cite{bay2006surf} in the YCbCr space as the image descriptor. For splicing forgeries, image features are calculated on block discrete cosine transform (DCT) in \cite{lin2011integrated}. \cite{muhammad2014image} identifies copy-move and splicing with a joint method using the local binary pattern descriptor. 
CFA based methods analyze camera internal filter patterns, assuming CFA structures in the tampered images have been damaged.  For example, \cite{Goljan2015CFAawareFF} detects the presence of CFA in images to identify the forgery. 

Recently, image noise residuals, obtained using high-pass filters, have shown increasing significance, especially for steganalysis rich model (SRM) \cite{Fridrich:2012:RMS:2713616.2713699}. Under the assumption that splicing and host images are characterized by different feature parameters, \cite{CozzolinoPV15} localizes splicing regions based on local image residuals. \cite{rao2016deep} uses the SRM filters as the bottom layer of CNNs to boost the forgery detection accuracy. \cite{zhou2018learning} combines noise residuals with RGB features via bilinear pooling. 
In fact, noise residuals are more significant than RGB contents for image forgery detection and three SRM filtering kernels are sufficient to determine the presence of the forgery. And what is more, residual kernels can be further optimized through iterative training on the basis of the SRM filters.

 
 \section{The Proposed Framework}
 
 \subsection{Image Fraud Identification}
 
\begin{figure*}[t]
 	\vskip 0.2in
 	\begin{center}
 		\centerline{\includegraphics[width=2\columnwidth]{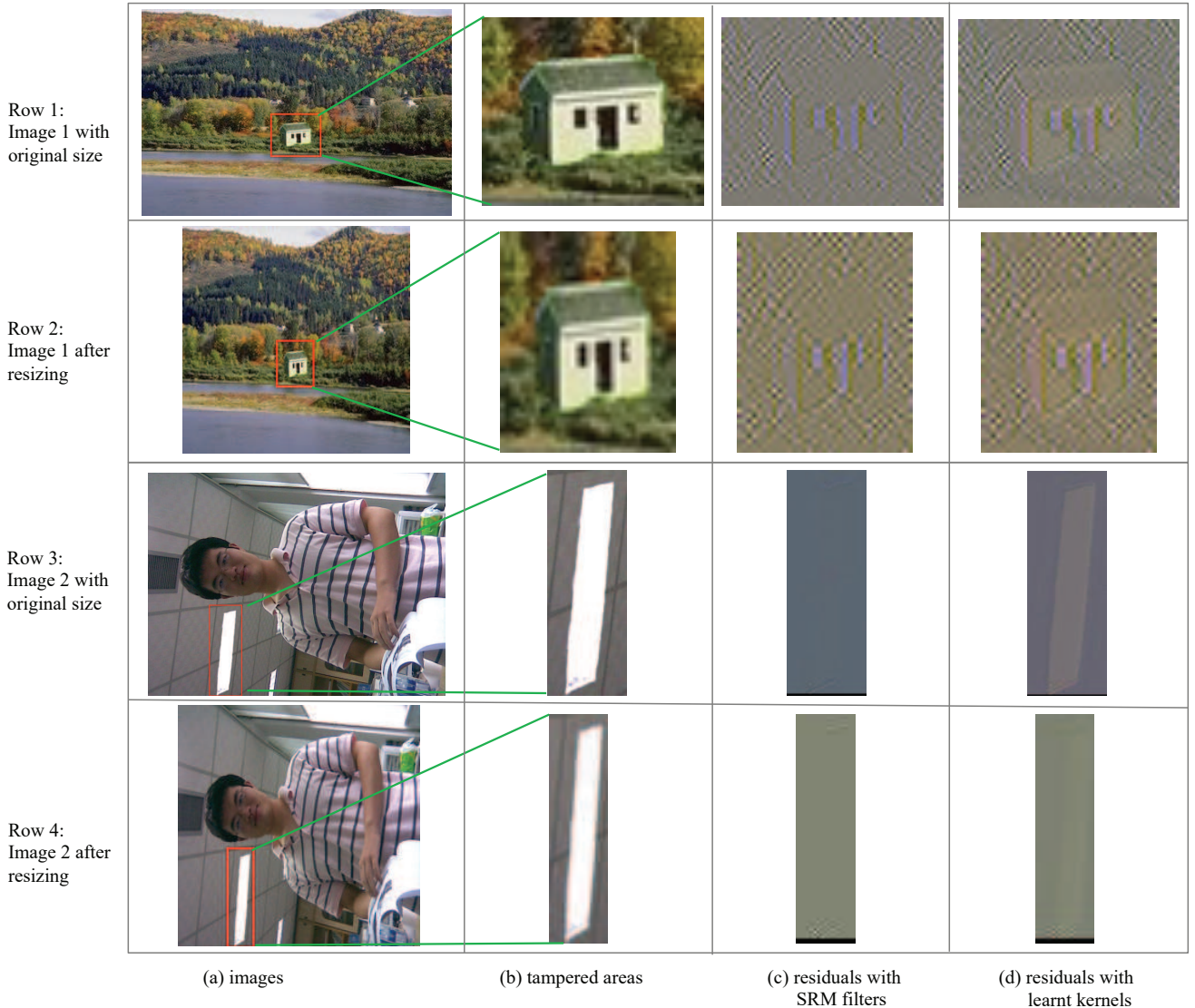}}
 		\caption{The effect of resizing on tampered areas. The \textit{1st} and \textit{3rd} rows are two images with original sizes: 384$\times$256 and 480$\times$640, where the red boxes represent the tampered areas respectively with the copy-move and splicing manipulations. When they were shrunk to 224$\times$224 (\textit{2nd} and \textit{4th} rows), the tampered areas would be resized and thus the noise residuals (\textit{3rd} and \textit{4th} columns) became poor distinctly. The last column shows that the residuals obtained with the learnt residual kernels are significant on tampered areas for original images. Best viewed in color.}
 		\label{fig:resize}
 	\end{center}
 	\vskip -0.2in
 \end{figure*}

Different from image forgery detection, image fraud identification aims to distinguish fraudulent images from genuine ones. As a consequence, we treat it as an image binary classification problem on the basis of CNNs.
The key to classification is to extract image features and then train a classifier with them. Based on recent advances on image forgery detecion, we expect to use image noise residuals to identify forged images. 
However, unfortunately, the microstructure in image noise residuals is vulnerable to the image resizing operation. The effect of resizing on tampered areas is especially substantial, which can be observed from Figure~\ref{fig:resize}. The 1st and 3rd rows show the original images of size $384\times 256$ and $480\times 640$ respectively, where tampered regions are marked out with red boxes. When the images are rescaled to $224\times 224$, as shown in the 2nd and 4th rows, the structures of noise residuals will be corrupted, especially for small tampered regions. The last two columns present the residuals obtained separately with the SRM filters and our learnt kernels. 
 
Considering the devastating impact of the resizing operation on image noise residuals, we propose a framework for image fraud identification through sequentially concatenating noise residual layer, feature extraction layers, SPP and softmax, as shown in Figure~\ref{fig:framework}. The noise residual layer is placed before networks and used to learn residual kernels, which will be optimized with an alternate training strategy (Section \ref{sec:train3}). This layer is in essence a convolutional operation as well and can be removed from the framework for general-purposed tasks. 
Feature extraction layers can be any backbone of prevailing CNNs such as VGG. SPP is added on the top of the last convolutional layer to ensure that the vector length of extracted features is uniform for arbitrary-sized images. 
Softmax layer is the last layer and resides at the end of SPP for classification.

In our implementation for training, each image is respectively fed to the network to deal with but model parameters are not updated until multiple gradients are accumulated. This optimization process, called pseudo-batch gradient descent, can guarantee that arbitrary-sized images are trainable on the original image scales when combining together with SPP. 
It is arbitrary-sized image training that preserves the fine microstructures in the noise residuals and thus results in better fraud identification performance.

 \subsection{Pseudo-batch Gradient Descent} \label{sec:pbgu} 
In this work, the loss function is defined as the cross-entropy loss $J(\theta)$ with regards to model parameters $\theta \in R^d$.
To optimize neural networks, gradient descent algorithms generally compute the loss $J_B(\theta)$ on a batch $B$ through summing the loss $J_{x_i, y_i}(\theta)$  on each example of this batch, 
 \begin{equation} 
 \label{eq:mini-batch-loss}
 J_B(\theta) = \alpha \sum_{x^i, y^i \in B} J_{ x^i, y^i}(\theta).
 \end{equation}
Here $\alpha$ is a scaling coefficient, $x^i$ denotes an input image and $y^i$ its corresponding label in a batch.
And model parameters are updated in the following way:
\begin{equation} 
\label{eq:mini-batch-update}
\theta \leftarrow \theta - \eta \nabla_\theta J_B(\theta),
\end{equation}
where $\eta$ is the learning rate.

The dilemma one may face in the practical implementation is that uniform tensors used in existing deep learning libraries cannot simultaneously handle images of varied sizes in a batch if without resizing or cropping beforehand. 
As a matter of fact, the mini-batch specified in these libraries involves both the images for computing gradients with tensors and those for parameter updates. 
To achieve arbitrary-sized image training, we
separate the batch into two independent ones in our implementation, namely the input batch $B_{n_i}$ of size $n_i$ and the update batch $B_{n_u}$ of size $n_u$. That is, our method
calculates the gradients for $n_i$ examples on $B_{n_i}$ at the same time for a tensor, but only updates the model coefficients after aggregating $n_u$ examples on $B_{n_u}$ at each iteration.
In this case, the parameter update process can be reformulated as:
 \begin{equation}
 \label{eq:pbgu}
 \theta \leftarrow \theta - \eta \sum_{B_{n_i}\subseteq B_{n_u}} \nabla_\theta J_{B_{n_i}}(\theta),
 \end{equation}
where $B_{n_i}$ is a subset of $B_{n_u}$.
Since the input batch $B_{n_i}$ works only for computing the gradients, it is actually a pseudo-batch. The real batch for parameter updates is $B_{n_u}$. 
Since $n_i$ is often specified as a variable to the mini-batch size, we call our implementation method as pseudo-batch gradient descent (PBGD) in this work.

In practice, there are two parameters regarding batch size, $n_i$ and $n_u$, to be preset in our method. For simplicity, $n_u$ can be initialized with a constant integer. It is worth noting that $n_u$ can be adaptively evaluated as well during training. For instance, denote $N$ as the average pixel number of an image in the training set, and $w_{x_i}, h_{x_i}$ as the width and height of the \textit{i}-th image $x_i$. 
For training images with different scales, $n_u$ can be described as the minimal number of images whose total pixels are not less than $N$,
and worked out as follows,
$$n_u= \lfloor n \rfloor,$$
$$s.t. \sum_{i=1}^{n} (w_{x_i} \times h_{x_i})\geq N$$
where $\lfloor\cdot \rfloor$ denotes the lower limit operation.

As discussed in Section \ref{sec:gd}, there is a close relationship between PBGD with BGD, SGD, and MBGD.
In practical applications, when $n_u$ is the same as $n_i$, PBGD turns into MBGD. It becomes SGD when both $n_u$ and $n_i$ are 1, and degenerates to BGD when $n_u$ equals the amount of all training examples.


%

 \subsection{Arbitrary-sized Image Training} \label{sec:arb}
Arbitrary-sized image training with CNNs is of great significance for image fraud identification as image resizing will damage potential noise residuals of primitive images. 
However, there are two challenges in arbitrary-sized image training. 

On one hand, the last layer, softmax, in CNNs generally requires fixed-length vectors as input. But if prohibiting the cropping or scaling operations, the length of extracted feature vectors will vary with the input image size. To resolve this problem, we place a SPP layer between the feature extraction layer and the softmax layer. SPP can fix the output feature size and finally generate fixed-length feature vectors from arbitrary-sized input images. 

On the other hand, the mini-batch size we can change is actually used for uniform tensors in deep learning libraries. If images are various in size, this batch size can only be set 1 and thus the gradient descent algorithm amounts to SGD. That will lead to the low converging stability. So we use the proposed PBGD method and rewrite the standard library codes for implemention in PyTorch. At this time, the input batch size $n_i$ is assigned 1 for arbitrary-sized image training.

 
 \subsection{Residual Kernel Learning} \label{sec:train3}
\begin{figure}[t]
  	\vskip 0.2in
  	\begin{center}
  		\centerline{\includegraphics[width=0.8\columnwidth]{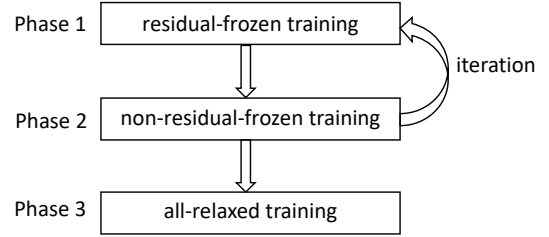}}
  		\caption{3-phase alternate training strategy.}
  		\label{fig:3phase}
  	\end{center}
  	\vskip -0.2in
  \end{figure}

 Image noise residuals are critical to image fraud identification due to its meaningful microstructure within the local regions \cite{Cozzolino:2017}. Specifically, under copy-move forgeries, microstructures show discontinuities and structural damages. Splicing forgeries are easy to be identified when detecting the mixture of different residual patterns due to the fact that some noise residuals exist specifically in an individual image. Various kernels have been designated to extract image noise residuals, e.g. SRM \cite{Fridrich:2012:RMS:2713616.2713699}, and they are also learnable in deep neural networks \cite{rao2016deep}. 
 The residual kernel layers usually locate in front of feature extraction layers and have weak gradients for updates. Hence, a 3-phase alternate training schedule is designed and utilized for learning residual kernels in this work, as illustrated in Figure \ref{fig:3phase}. 
In the first phase, the parameters of residual kernels are frozen and the other layers are trained. Then we only train the residual layer while fixing the remaining. Finally, all parameters are relaxed and trained together after alternately iterating the first two phases until convergence.

\begin{figure*}[t]
 	\vskip 0.2in
 	\begin{center}
 		\centerline{\includegraphics[width=1.5\columnwidth]{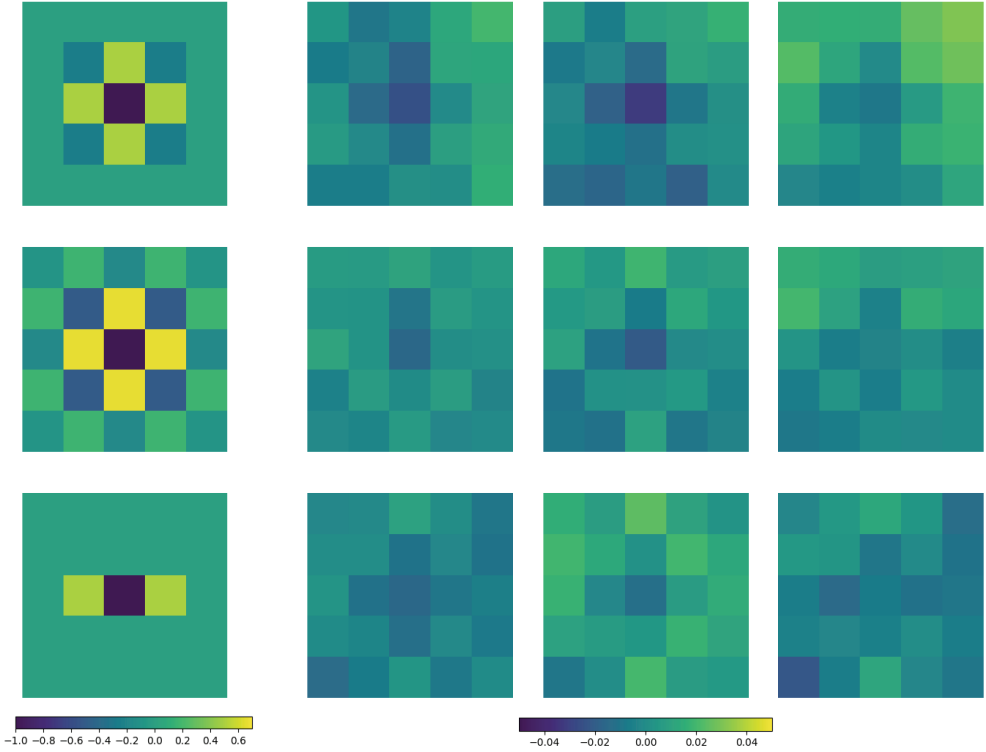}}
 		\caption{The learnable residual kernels. The left column lists three SRM filters used for initialization. The right three columns illustrate the learnt kernels for \textit{r}, \textit{g}, and \textit{b} channels, respectively.  Best viewed in color.}
 		\label{fig:kernel}
 	\end{center}
 	\vskip -0.2in
 \end{figure*}

In practical applications, our residual kernels are initialized by the three most significant SRM filters in the first column of Figure~\ref{fig:kernel}. In image fraud identification, our 3-phase alternate training makes a fine distinction between the initial kernels and the learnt ones, as presented in Figure~\ref{fig:kernel}. The learnt kernels are different as well for three color channels, \textit{r}, \textit{g}, and \textit{b}.  Let us get back to the two image examples in Figure~\ref{fig:resize}. It is easily observed that
the noise residuals (the 4th column), obtained with the learnt kernels, look clearer than the counterparts (the 3rd column) with the original SRM filters. More significantly, the structures of the residuals the learnt kernels extract for original images are more normal on tampered regions than those after resizing.
And the considerably improved accuracy in our experiments of Section \ref{exp:ifi} also corroborate our assumption that the well trained residual kernels are more beneficial to image fraud identification.
 
 \section{Experiments}
 \subsection{Experimental Setup}

 In our experiments, we used the following four standard datasets: CASIA v1.0 and v2.0 \cite{dong2011casia}, Columbia \cite{Ng2004}, and COVER\cite{wen2016coverage}, as listed in Table \ref{tab:data}.
 The \textbf{CASIA} v1.0 dataset contains 800 authentic and 921 forged
 color images 459 of which are copy-move forged, and the
 remaining are spliced. Different geometric transformations, such
 as scaling and rotating, have been applied on the forged
 images. All the images have a size of 384$\times$256 in the JPEG format. The CASIA 2.0 includes 7,491 authentic and 5,123 forged color images with size ranging from 240$\times$60 to 900$\times$600 in the JPEG, BMP or TIFF formats. The \textbf{Columbia} image
 database consists of 183 authentic and 180 spliced images in either TIFF or BMP formats. The images are uncompressed and vary in size from 757x568 to 1152x768.
 \textbf{COVER} is composed of 100 original–forged image pairs with image size of 400$\times$486, and only focuses on the copy-move forgery. This dataset is challenging since it covers similar objects at the pasted regions to conceal the tampering artifacts. 
 
 These available benchmark datasets are usually small-scale and directly training CNNs with them is prone to model overfitting. To eliminate the problem, we collected a set of data comprising of 84k forged images from the PS-Battle dataset \cite{heller2018ps} , 11,833 forged ones from the Zhou dataset \cite{zhou2018generate}, and 100k authentic ones from the Place365 dataset \cite{zhou2017places} to increase the diversity of training examples. In this collection, the image side length ranges from below 100 pixels through 20,000 pixels. For forged images, the tampered region sizes vary a lot and several tampering techniques were applied, e.g. copy-move, splicing and partial removal. This diversity in the image resolutions, the tampered region sizes and the tampering techniques helps pre-train a more powerful model with the capability of better representation and strong generalization. 
  
\begin{figure}[!th]
  	\vskip 0.2in
  	\begin{center}
  		\centerline{\includegraphics[width=\columnwidth]{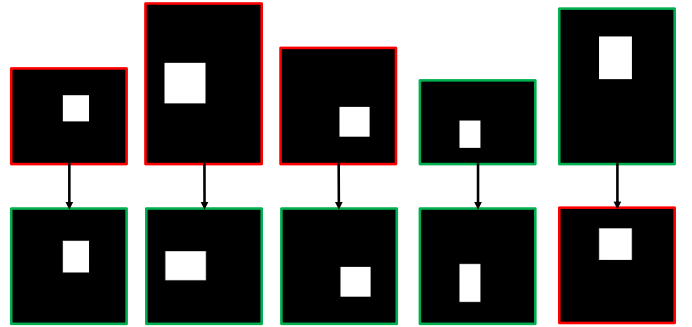}}
  		\caption{Some examples of the Square dataset. The first row shows the original images from the dataset and the second row is the counterparts after resizing to a fixed size. The images with red borders are with class label 'square' and those with green are 'non-square'. The semantic categories are messed after the resizing operation. Best viewed in color.}
  		\label{fig:square}
  	\end{center}
  	\vskip -0.2in
  \end{figure}

 In addition,
 a toy dataset, named \textbf{Square}, was also synthesized and studied for analyzing the impact of resizing on the classification performance. It comprises of 12,000 images from 2 classes, \textit{square} and \textit{non-square}. The image aspect ratio in this dataset ranges from 0.6 to 1.67 and some examples are displayed in the top row of Figure~\ref{fig:square}.

 \newcommand{\tabincell}[2]{\begin{tabular}{@{}#1@{}}#2\end{tabular}}  

 \begin{table}[t]
 	\caption{Summary of four benchmark datasets. There are only two
categories in each dataset: authentic (au.) category and forged (fo.)
category.}
 	\label{tab:data}
 	\vskip 0.15in
 	\begin{center}
 		\begin{tabular}{lccccc}
 			\toprule
 			 \multirow{2}{*}{dataset} &\multicolumn{2}{c}{amount} &tampering&\multirow{2}{*}{size} &\multirow{2}{*}{format}\\
 			  \cline{2-3}
 			 &au. &fo.&type&&\\
 			\midrule
 			CASIA v1.0 & 800& 921&\tabincell{c}{copy-move \\ splicing}&$384\times 256$& JPEG\\
 			\hline
 			CASIA v2.0 & 7,491& 5,123&\tabincell{c}{copy-move \\ splicing} &\tabincell{c}{$240\times 60\sim$\\ $900\times 600$}&\tabincell{c}{JPEG\\BMP\\ TIFF}\\
 			 \hline
			Columbia & 183 &180 &splicing&\tabincell{c}{$757\times 568\sim$\\ $1152\times 768$} &\tabincell{c}{BMP\\ TIFF}\\			 		
  			\hline
 			COVER& 100& 100&copy-move&$400\times 486$&TIFF\\
 			\bottomrule
 		\end{tabular}
 	\end{center}
 	\vskip -0.1in
 \end{table}
 
The accuracy acts as the metric for performance evaluation, which can be formulated as follows:
 \begin{equation}
 \label{eq:acc}
 accuracy = \frac{TP+TN}{TP+TN+FP+FN},
 \end{equation}
where $TP$, $TN$, $FP$ and $FN$ represent the number of \textit{true positive, true negative, false positive, and false negative} examples respectively.

 All the experiments were performed on a server with Intel(R) Core(TM) i7-7700 CPU @ 3.60GHz, 16G memory and a NVIDIA GeForce TITAN X (Pascal) GPU.
  
\subsection{Analysis on Arbitrary-sized Image Training}
To verify the effect of the proposed framework on arbitrary-sized image training, we conducted tests involving image classification on the Square dataset.
In these experiments, we removed the residual layer from our framework, 
and adopted the backbone of VGG16 \cite{Simonyan2014VeryDC} as feature extraction layers. For the purpose of comparison, the original \textit{VGG16} using the resizing operation and its modified version \textit{VGG16-p} with PBGD were tested on the same dataset as well. In order to distinguish our method from others by name, we named it VGG16-sp in this case.
The notable difference between VGG16-sp and the other two is that the former can accept arbitrary-sized images as input, but the latters require the resizing operation to produce images of fixed size (224$\times$224).

In our experiments, model parameters were initialized with the weights pre-trained on ImageNet. The learning rate in PBGD was initially set to 0.0001, and divided by 10 when the validation accuracy did not improve in consecutive 5 epochs. All models were trained for 10 epochs with update batch sizes ($n_u$) of 4 and 8, and the SPP layers were constructed with 3-scale outputs, 1$\times$1, 2$\times$2 and 4$\times$4. To no one's surprise, the VGG16-sp performed 100 percent correctly on this synthetic dataset, much better than both VGG16 and VGG16-p, as shown in Table~\ref{tab:pbgu_general}.

 
 \begin{table}[!h]
 	\caption{Classification on the Square dataset with different strategies of scaling and gradient updates.}
 	\label{tab:pbgu_general}
 	\vskip 0.15in
 	\begin{center}
 		\begin{tabular}{lcc}
 			\toprule
 			 \multirow{2}{*}{network}&\multicolumn{2}{c}{batch size ($n_u$)}  \\
 			 \cline{2-3}
 			&$n_u=4$  &$n_u=8$ \\
 			\midrule
 			VGG16& 0.5257& 0.5712 \\
 			VGG16-p& 0.6523& 0.5013\\
 			VGG16-sp& \textbf{1.0}& \textbf{1.0} \\
 			\bottomrule
 		\end{tabular}
 	\end{center}
 	\vskip -0.1in
 \end{table}

From the experimental results, we observed that VGG16 and VGG16-p almost resulted in random guesses, mainly due to the resizing operation altering the original semantic context in images. 
Some examples are demonstrated in Figure~\ref{fig:square}. When the original images (top row) were resized to a fixed size (bottom row), the 'square'
shape embedded in the images (first 3 columns) changed to the general rectangle 'non-square', and contrarily the non-square in the last column deformed to the square. However, the arbitrary-sized image training surely keeps unchanged the semantic features in images through PBGD plus SPP. SPP sets free the CNNs from the constraint of the fixed size and PBGD resolves the discrepancy between the parameter update and the tensor design in deep learning libraries.  
This highlights the practical significance of the proposed framework to image classification.

 

 \subsection{Analysis on Image Fraud Identification}\label{exp:ifi}

 Our empirical studies also revealed that VGG16 performed the best if we set VGG13, VGG16, VGG19 and AlexNet respectively as the backbone for feature extraction layers in the proposed framework.
 As a result, we selected VGG16 as the backbone for image fraud identification in the following experiments. 
 Our collected data was randomly split into two subsets, training (80\%) and validation (20\%), and used to pre-train a fraud identification model.
 
 \subsubsection{Improvement with Residual Kernel Learning}
 
 \begin{table}[t]
 	\caption{Validation accuracy with the alternate training strategy}
 	\label{tab:alternate}
 	\begin{center}
 		\begin{tabular}{llccc}
 			\toprule
 			\multirow{4}{*}{iteration}&no.&residual-frozen&non-residual-frozen \\
 			\cline{2-4}
 			&1&0.9634&0.9599 \\
	 		\cline{2-4}
 			& 2&0.9769&0.9791\\
	 		\cline{2-4}
 		   & 3&0.9816&0.9813 \\
 			\midrule
 		   all-relaxed &\multicolumn{3}{c}{\textbf{0.9903}} \\
 			\bottomrule
 		\end{tabular}
 	\end{center}
 \end{table}
 

Image fraud identification pays more attentions to noise residuals than common image classification tasks. 
In order to find better residual kernels, we executed the training on the collected data with the 3-phase alternate training strategy in the proposed framework. The residual kernels were initialized with the SRM filters \cite{Fridrich:2012:RMS:2713616.2713699} and continued to be optimized until they were able to better describe noise residuals. In the following, our method is named VGG16-spk for the convenience of memory. As shown in Table~\ref{tab:alternate}, the learnt residual kernels can obviously improve the validation accuracy, from 0.9634 to 0.9813, after three alternate iterations. When all model parameters were relaxed for further training at the last phase, the accuracy increased to 0.9903. Nine resulting kernels are shown in the right three columns of Figure~\ref{fig:kernel}.

\begin{figure}[t]
 	\vskip 0.2in
 	\begin{center}
 		\centerline{\includegraphics[width=1\columnwidth]{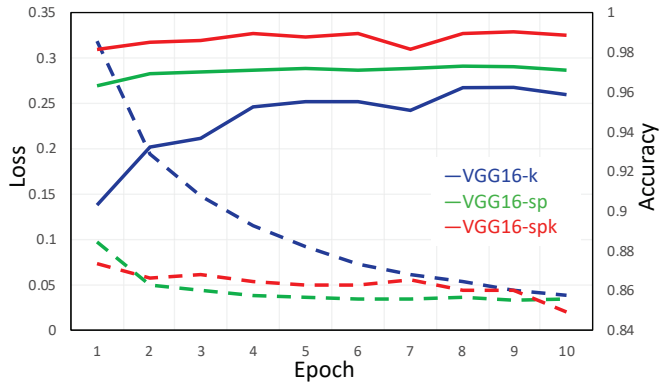}}
 		\caption{The validation accuracy and convergence rate of VGG16-k, VGG16-sp and VGG16-spk. The dash and solid lines respectively denote the change of training loss and validation accuracy with epochs. Best viewed in color.}
 		\label{fig:converge}
 	\end{center}
 	\vskip -0.2in
 \end{figure}

For fair comparison, two baselines named VGG16-k and VGG16-sp were trained on the collected data respectively. 
VGG16-k was based on VGG-16, but rather filtered input images with learnt residual kernels after resizing.  
Figure~\ref{fig:converge} illustrated the validation accuracy of these three methods. The results unveiled the significance of residual kernel learning to the proposed framework, where VGG16-spk achieved better accuracy than VGG16-sp. It is also understandable that VGG16-k performed worst as the resizing operation has already damaged the noise residuals.
 
 To further validate the effect of the learnt kernels, we also conducted tests on CASIA v2.0 with different networks. Here, the CASIA v2.0 dataset was split into 6 disjoint groups five of which were used for finetuning and the left for test. Another method, called VGG16-spf, was also tested for comparative analysis. This method is basically identical to VGG16-spk except that the residual kernels kept unchanged once initializing with the SRM filters.
 As seen in Table~\ref{tab:small_casia2}, VGG16-spk had the best average accuracy of 0.9947 among all methods. More significantly, using the residual kernels learnt with the alternate training strategy, we improved the accuracy over 1\% than VGG16-spf using the original SRM filters, nearly 3\% than VGG16-sp only using the RGB content, and almost 8\% than VGG16-k combining the learnt residual kernels with the resizing operation.
 
 \begin{table}[!h]
 	\caption{Accuracy of different methods on CASIA v2.0 and its subsets involving different tampered region sizes.}
 	\label{tab:small_casia2}
 	\begin{center}
 		\begin{tabular}{lcccc}
 			\toprule
 			\multirow{2}{*}{method}&\multicolumn{3}{c}{tampered region} &\multirow{2}{*}{CASIA v2.0}\\
 			\cline{2-4}
 			&small&medium&large &\\
 			\midrule
 			VGG16-k  & 0.9122& 0.9506& 0.9682& 0.9176\\ 
 			VGG16-sp  & 0.9615& 0.9741& 1.0& 0.9649\\
 			VGG16-spf  & 0.9831& 0.9803& 1.0&0.9829\\
 			VGG16-spk  & \textbf{0.9953}& \textbf{0.9924}& \textbf{1.0}&\textbf{0.9947}\\
 			\bottomrule
 		\end{tabular}
 	\end{center}
 \end{table}
 
 \subsubsection{Superiority on Small Tampered Regions}
 Small tampered regions often happened to fraudulent images, especially for loan or insurance frauds in Internet finance business.
As stated previously, the main advantage of arbitrary-sized image training is that it can guarantee that the fine features in small regions will not be changed during training. Specifically, to analyze this superiority in image fraud identification, we classified the CASIA v2.0 dataset into three subsets, \textit{large}, \textit{medium}, and \textit{small}, in terms of the tampered region size. A tampered region is generally defined as \textit{small} if its area accounts for below 20\% of the whole image area, \textit{large} if over 50\%, and \textit{medium} otherwise.
 
Separately computing the accuracy of three subsets, we can find that the arbitrary-sized image training strategy, adopted in the bottom three methods of Table~\ref{tab:small_casia2}, was obviously more beneficial to small tampered regions than that of resizing images to a fixed size in the VGG16-k method. The proposed framework VGG16-spk gained over 8\% in the \textit{small} subset, 
which is the superiority of residual kernel learning and arbitrary-sized image training towards image fraud identification.
 


 \subsubsection{Generalization to New Dataset}

COVER is a small dataset and therefore is unsuitable for solely training deep models. Here we conducted tests on the COVER dataset directly with the finetuned model on CASIA v2.0, so as to evaluate the generalization capability of the proposed method. Traditional methods, SIFT\cite{SIFT2003}, SURF\cite{SURF2010}, and Dense-Field \cite{Dense2015}, were selected as the baselines, and the corresponding results are available in \cite{wen2016coverage}. For fair comparison, from each original-forged pair, we randomly picked one to predict. As shown in Table~\ref{tab:general}, the proposed method, VGG16-spk, outperformed the cutting-edge \textit{Dense-Field} by nearly 5\% in accuracy. It strongly corroborates a fact that our method has better generalization ability for unknown images.
 
  \begin{table}[!h]
  	\caption{Accuracy on the COVER dataset}
  	\label{tab:general}
  	\begin{center}
  		\begin{tabular}{lcccc}
  			\toprule
  			SIFT&SURF&Dense-Field&VGG16-spk\\
  			\midrule
  			0.5050&0.5860&0.7180  & \textbf{0.7650}\\
  			\bottomrule
  		\end{tabular}
  	\end{center}
  \end{table}
  
 \subsubsection{Comparison with Other Methods }

For comprehensive analysis, we tested the proposed method on three standard datasets, Columbia, CASIA v1.0 and v2.0, and compared it with four state-of-the-art methods,
Rao \cite{rao2016deep}, Muhammad \cite{muhammad2014image}, Shi \cite{Shi2007ANI}, and Zhao \cite{Zhao2012OptimalCC}, in this experiment.
From Table~\ref{tab:compare}, we observed the proposed method achieved best results on all datasets. It is worth noting that our method is more suitable for datasets with various (CASIA v2.0) or large (Columbia) image sizes. For image sets with fixed size and small resolution, the performance superiority is not that much. In particular, the proposed method only improved 0.27 percent in accuracy on CASIA v1.0, but the gain can reach 1.64 percent on CASIA v2.0 and 3.03 percent on Columbia over the SOTA results.
 
 \begin{table}[!h]
 	\caption{Comparison with other methods on three datasets.}
 	\label{tab:compare}
 	\begin{center}
 		\begin{tabular}{lccc}
 			\toprule
 			method &  CASIA v1.0&CASIA v2.0&Columbia  \\
 			\midrule
 			Rao&0.9804 &0.9783&-\\
 			Muhammad& 0.9489&0.9733&0.9639\\
 			Shi& -&0.8486&-\\
 			Zhao&-&-&0.9314\\
 			VGG16-spk& \textbf{0.9831}&\textbf{0.9947}&\textbf{0.9942}\\
 			\bottomrule
 		\end{tabular}
 	\end{center}
 \end{table}

 \subsection{Computational Performance} \label{sec:exp_time}
To thoroughly analyze the computational performance of the proposed method, we examined the training and inference time cost separately.

During the 3-phase alternate training, the residual kernel learning is very fast as there are few weights to optimize in residual kernels when non-residual layers are frozen. As a consequence, its computational cost can be ignored in the whole training period and we will focus more on computation overhead of the arbitrary-sized image training.

In this experiment, we compared our training strategy VGG16-sp with several other methods, VGG16, VGG16-bn, VGG16-gn and VGG16-s. All methods are based on the VGG16 backbone, where VGG16 is the original version, VGG16-bn and VGG16-gn respectively adopt batch normalization and group normalization, VGG16-s means to add a SPP layer to the backbone, and VGG16-sp updates parameters with PBGD besides adding the SPP layer. 
In this case, the mini-batch size of VGG16-s was set to 1 and the other mini-batch sizes were 16. As to VGG16-sp, the input batch size $n_i$ and update batch size $n_u$ were respectively 1 and 16. Each method ran 10 epochs on the same dataset and its average training time was computed in terms of epochs. As illustrated in Table~\ref{tab:train_time}, VGG16-sp is more efficient than VGG16-s through avoiding frequent back-propagation, but slower than the others because the arbitrary-sized image training can not take advantage of the tensor structure in deep learning libraries to quickly calculate multiple gradients once.

 \begin{table}[!h]
 	\caption{Comparison of the average training time per epoch with different methods.}
 	\label{tab:train_time}
 	\vskip 0.15in
 	\begin{center}
 		\begin{tabular}{lccccc}
 			\toprule
 			\multirow{2}{*}{method} & \multicolumn{5}{c}{VGG16} \\
 			\cline{2-6}
 			&/&-bn&-gn&-sp &-s \\
 			\midrule
 			time(s)   &19.2& 21.3& 25.6& 55.7&68.8\\
 			\bottomrule
 		\end{tabular}
 	\end{center}
 	\vskip -0.1in
 \end{table}
 
Next we studied the impact of the update batch size $n_u$ on computation overhead in the arbitrary-sized image training. It is well known that the overall training time is mainly composed of two components, one for accumulating prediction errors and the other for model updates. The input batch size $n_i$ was unchanged (always 1) in the arbitrary-sized image training, so the overall training time for an epoch is primarily dependent on the model update time. When $n_u$ is small, the updating process has to be frequently called and the computational cost will be high. With the increase of $n_u$, the overhead should decrease gradually. However, the updates require the additional complexity of computing gradients across $n_u$ training examples. Accordingly, model updates, and in turn training speed, may become slow for large $n_u$. 

Our experiments confirmed the conclusion obtained from the above analysis.
Table~\ref{tab:different-batch} presents the average training time per epoch with regards to $n_u$ varying from 1 to 128. The training cost decreased quickly for smaller $n_u$ but stably stayed around 11 seconds when $n_u$ became no less than 16. As a result, it is reasonable and practical to set the update batch size to 16 in real applications. Moreover, our experiments in Figure \ref{fig:converge} demonstrated that VGG-spk with $n_u$ of 16 can converge much faster than VGG-k and basically as quickly as VGG-sp.

 \begin{table}[t]
 	\caption{The average training time (s) of various update batch sizes ($n_u$) per epoch.}
 	\label{tab:different-batch}
 	\vskip 0.15in
 	\begin{center}
 		\begin{tabular}{l|cccccccc}
 			\toprule
 			$n_u$ &1 &2&4&8 &16&32&64&128\\
 			\hline
 			time &17.2 &14.0&12.4&11.7 &11.4&11.3&11.3&11.3\\
 			\bottomrule
 		\end{tabular}
 	\end{center}
 	\vskip -0.1in
 \end{table}
 
 Since the proposed method can take arbitrary-sized image as input, its inference time should be closely correlated with the input image size. The inference time involving different image resolutions from 224$\times$224 through 1920$\times$1920 is shown in Table~\ref{tab:different-resolution}. It can be observed that the inference time increased roughly linearly with the image resolution and reached to 168.4 \textit{ms} for the image of 1920$\times$1920, which was quick enough and thus acceptable to real-time image fraud identification. 
 
 
 
  \begin{table}[!h]
  	\caption{The inference time per image of different input resolutions}
  	\label{tab:different-resolution}
  	\vskip 0.15in
  	\begin{center}
  		\begin{tabular}{l|cccc}
  			\toprule
  			resolution & 224$\times$224&512$\times$512&1024$\times$1024&	1920$\times$1920\\
  			\hline
  			time (ms)& 4.3& 14.6&55.1& 168.4\\
  			\bottomrule
  		\end{tabular}
  	\end{center}
  	\vskip -0.1in
  \end{table}
 
 
 \section{Conclusion}
 In this work, for the purpose of arbitrary-sized image training, we proposed pseudo-batch gradient descent (PBGD) and redesigned the key modules in deep learning libraries to accumulate gradients sequentially on the update batch.
To improve the performance on image fraud identification, we conceived a 3-phase alternate training strategy to learn optimal residual kernels. 
Leveraging residual kernel learning and PBGD, the proposed framework effectively extracted and preserved image noise residuals in original images. Extensive experimental evaluation demonstrated that the proposed method outperformed the commonly used fixed-sized training strategy for image classification thanks to its
salient strength in preserving semantic contexts of images during training.
More significantly, for image fraud identification, our method achieved the state-of-the-art results as well, especially for images with small tampered regions or unseen images with different tampering distributions.


%

%
%


\ifCLASSOPTIONcaptionsoff
  \newpage
\fi



\bibliographystyle{IEEEtran}
\bibliography{main}

\end{document}